\documentclass[10pt,twocolumn,letterpaper]{article}

\usepackage{ijcb}
\usepackage{times}
\usepackage{epsfig}
\usepackage{graphicx}
\usepackage{amsmath}
\usepackage{amssymb}
\usepackage[norule,symbol,perpage]{footmisc}

\usepackage{subfigure}

\ijcbfinalcopy % *** Uncomment this line for the final submission

\ifijcbfinal\fi

\makeatletter
\def\ps@IEEEtitlepagestyle{
\def\@oddfoot{\mycopyrightnotice}
\def\@evenfoot{}
}
\def\mycopyrightnotice{
{\hfill \footnotesize 978-1-6654-3780-6/21/\$31.00 \copyright 2021 IEEE\hfill}
}
\makeatother

\begin{document}

%%%%%%%%% TITLE
\title{An End-to-End Autofocus Camera for Iris on the Move}

\author{
Leyuan Wang, Kunbo Zhang, Yunlong Wang, Zhenan Sun\\
School of Artificial Intelligence, UCAS\\
Center for Research on Intelligent Perception and Computing\\
National Laboratory of Pattern Recognition, CASIA\\
{\tt\small \{leyuan.wang, yunlong.wang\}@cripac.ia.ac.cn}, {\tt\small kunbo.zhang@ia.ac.cn}, {\tt\small znsun@nlpr.ia.ac.cn}
}

\maketitle
\thispagestyle{empty}

%%%%%%%%% ABSTRACT
\begin{abstract}
   For distant iris recognition, a long focal length lens is generally used to ensure the resolution of iris images, which reduces the depth of field and leads to potential defocus blur. To accommodate users at different distances, it is necessary to control focus quickly and accurately. While for users in motion, it is expected to maintain the correct focus on the iris area continuously. In this paper, we introduced a novel rapid autofocus camera for active refocusing of the iris area of the moving objects using a focus-tunable lens. Our end-to-end computational algorithm can predict the best focus position from one single blurred image and generate a lens diopter control signal automatically. This scene-based active manipulation method enables real-time focus tracking of the iris area of a moving object. We built a testing bench to collect real-world focal stacks for evaluation of the autofocus methods. Our camera has reached an autofocus speed of over 50 fps. The results demonstrate the advantages of our proposed camera for biometric perception in static and dynamic scenes. The code is available at https://github.com/Debatrix/AquulaCam.
\end{abstract}

\let\thefootnote\relax\footnotetext{\mycopyrightnotice}

%%%%%%%%% BODY TEXT
\section{Introduction}

\begin{figure}[h]
   \begin{center}
      \includegraphics[width=1\linewidth]{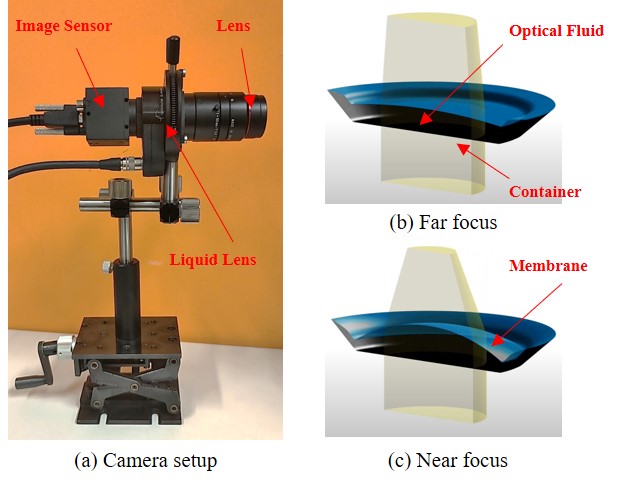}
   \end{center}
   \caption{Illustrations of camera setup (a) and focus tunable lens working principle.}
   \label{fig:AquulaCam}
\end{figure}

\begin{figure*}[h]
   \begin{center}
      \includegraphics[width=0.95\linewidth]{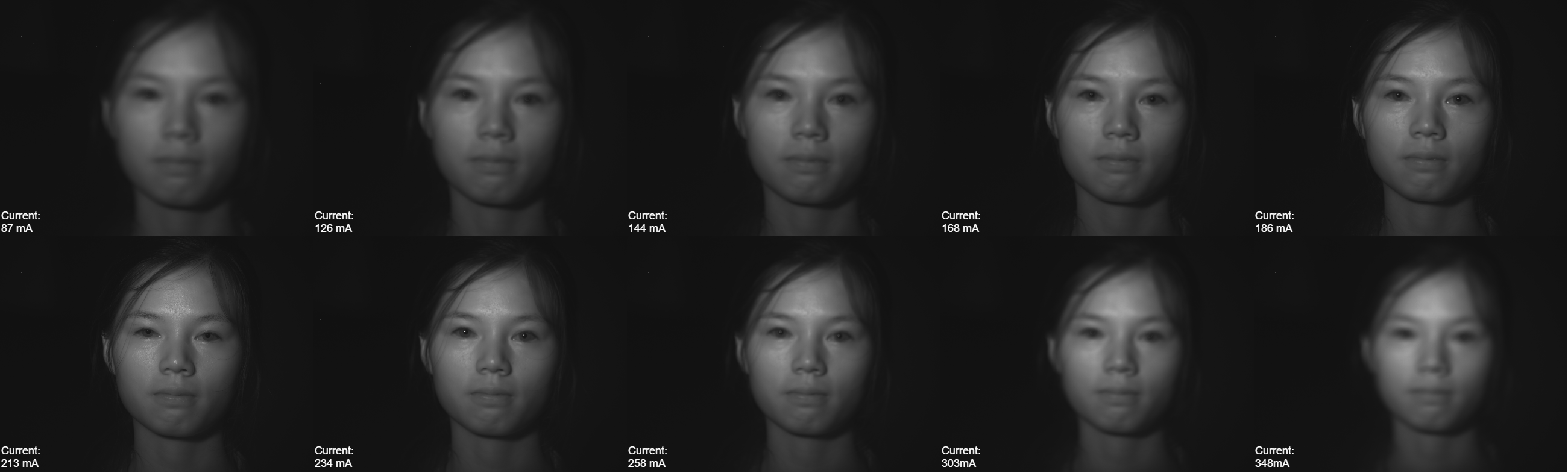}
   \end{center}
   \caption{A focal stack of face images shows the transition between blurry and clear using the focus-tunable lens.}
   \label{fig:focal_stack}
\end{figure*}

Defocus blurring is one of the essential quality impacts in iris recognition for long-range uncontrolled objects\cite{iso_29794,kalka2010estimating,Tabassi_Grother_Salamon_2011}. The lens of an iris imaging system with a long focal distance is introduced to meet the high-resolution requirement of an iris image, which sacrifices the field of depth (DoF)\cite{iom_2006,Nguyen_Fookes_Jillela_Sridharan_Ross_2017,Zhang_Shen_Wang_Sun_2020}. Uncontrolled subjects can quickly leave the range of DoF, making it difficult to acquire clear iris images. To this end, we have developed AquulaCam that can actively rapidly autofocus on human eyes and maintain a good focus on the moving object. This autofocus (AF) camera is also expected to solve the problems in other scenarios that require active autofocus, continuous tracking, and precise focus position, such as live sports broadcasts and mobile phone anti-shake.

Although digital image processing has made significant developments in recent years, recovering clear images from defocused images is still a considerable challenge\cite{brady2018parallel}. The best solution to this problem is to ensure that all captured images are in good focus through autofocus adjustment. It can realize in either hardware-based or algorithm-based methods. The hardware-based method has a relatively faster speed and can maintain a good focus position even while an object is moving. But this requires additional cost and is restricted by hardware limits. The algorithm-based method does not require complicated hardware design except for the basic camera itself, but the conventional adjustment steps are time-consuming to achieve satisfactory results.

Moreover, neither of these two methods can actively perceive specific objects in the scene. On the contrary, human beings can only rely on the eyes and brain to actively focus on particular things in the field of view, and can maintain the object in the ideal focus position when it is moving. Inspired by this biological mechanism, we proposed an AI-controlled camera named AquulaCam. The term Aquula represents the crystalline lens of an eyeball. Our AquulaCam uses an active object tracker based on deep learning, which actively perceives specific objects from only one single image and learns to generate the camera focus control signal in an end-to-end computational manner. Besides the rapid autofocus in a static scene, it is able to maintain the focus tracking on a moving object.

AquulaCam consists of an image sensor, a liquid lens, and an ordinary lens, as shown in Figure \ref{fig:AquulaCam}. The most important part is the liquid lens, which is shape-changing lenses based on a combination of optical fluids and a polymer membrane\cite{Optotunelens}. Applying an electric current to this shape-changing polymer lens, its optical power is adjusted within about 25 ms. This fast response and high precise focus control tunable optical element provide us the possibility for real-time focus tracking of a moving object.

To develop the control algorithms for AquulaCam, we first derived the relationship between the point spread function (PSF) of a blurred image and the camera's focal length. It is demonstrated the feasibility of employing a convolutional neural network (CNN) to generate lens diopter control signal straightforward. Then a representation of objects in blurred images and its detection method is designed for active perception. Finally, we built an active end-to-end model, collected a dataset containing 76 focus stacks like figure \ref{fig:focal_stack}, designed a virtual camera for offline training and testing, and investigated our model's performance of a virtual camera and real scenes.

The major contributions of this paper are:
\begin{itemize}
   \item \textbf{Rapid autofocus and continuous focus tracking for moving objects.} We demonstrate that the active tracking model based on deep learning can generate a lens diopter control signal from a single blurred image. The speed of our algorithm and the motion modeling capability enables it to maintain a good focus on the moving object.
   \item \textbf{An end-to-end active visual perception model.} Our algorithm can actively perceive the iris area without detection of the region of interest (ROI). It can also directly generate a lens diopter control signal without additional calibration and manipulation of the camera. Thereby constructing an end-to-end active tracking model which can predict the ideal focus position from the original image.
\end{itemize}

%-------------------------------------------------------------------------
\section{Related Work}
\label{sec:related}

\subsection{Object Tracking}

Object tracking aims to localize an object in a continuous image sequence given an initial bounding box in the first frame. Extensive researches have been conducted in both passive and active scenarios. Passive object tracking is assumed that the tracked object is always in the image scene, and there is no need to control the camera during tracking. Correlation filter based object tracking \cite{bolme2010visual}, \cite{henriques2014high} has achieved a success in real-time object tracking, and deep learning has also been successfully applied to object tracking \cite{li2019siamrpn++}, \cite{bertinetto2016fully} in recent years. Active tracking unifies the two subtasks of localizing object and camera control. Conventional solutions dealt with object tracking and camera control in separate components\cite{tilmonFoveaCamMEMSMirrorEnabled2020,denzler1994active,torkaman2012real}. \cite{tilmonFoveaCamMEMSMirrorEnabled2020} use an optional Kalman filter to estimate the current state of the MEMS mirror's 1D motion and the object locations, given a previous state and object locations and the desired control vector. Besides, there is tackling object tracking and camera control simultaneously through reinforcement learning \cite{luoEndtoendActiveObject2018, zhong2019ad}.

However, there exists not much research attention in the area of active tracking. And there is very little research on using active tracking to control the camera's autofocus and continuously maintain the moving object in a proper focus position. It is difficult to adjust the solution to perceive the moving object and adjust the camera focus separately. Our proposal completes the perception of moving objects and the control of imaging equipment in an end-to-end manner.

\subsection{Autofocus}

There are three main strategies for autofocus: active distance sensing, phase detection, and contrast maximization. Active distance sensing uses active sensors such as radar or structured light to detect the distance between the target and the camera and adjust the focus accordingly. Phase detection uses light field sensors to measure the disparity between the current focus setting and the in focus setting\cite{pengEnhancedCameraCapturing2016,sliwinskiSimpleModelOnSensor2013,chanEnhancementPhaseDetection2017}. Both active distance sensing and phase detection require additional hardware. They can determine the correct focus in a single time step and can be fast enough for moving objects. But in particular, unlike an external distance sensor, the light field sensor required for phase detection needs to share the optical path with the original imaging device, which is not easily modified. The active distance sensor's error is relatively large, and it may more than 10cm at a distance of greater than 3m. If the imaging device's depth of field is less than the error range of the active distance sensor, it will not be possible to focus. The method we propose is directly based on the captured image to focus, without additional hardware, and the accuracy is consistent with the accuracy of the imaging device.

Contrast maximization uses image quality measures to evaluate the focal quality and search for optimal focus. It is image-based and requires no hardware beyond the basic camera itself. Traditional systems apply a search strategy to maximize the quality measures. Popular strategies include Fibonacci search\cite{gendlin1982focusing}, rule-based search\cite{kehtarnavaz2003development} and hill-climbing search\cite{he2003modified}. Ideally, such strategies should use as few steps as possible. To reduce the number of search steps, some methods based on quality curve fitting are proposed\cite{yazdanfar2008simple,chiu2010efficient}. In recent years, a lot of auto-focus methods based on machine learning have also appeared\cite{park2008fast,chen2010passive,han2011novel}, but these methods still require a large number of hand-designed features and complex data processing. If the object moves, the image quality curve will also change, and the search space will change accordingly. Therefore, these methods cannot effectively focus on the moving object. Our proposed method can directly estimate the distance between the image and the best focus position from a single image and can model the object's movement to achieve continuous focus on the moving object.

%-------------------------------------------------------------------------
\section{Our Approach}
\label{sec:approach}

Our goal is to keep the precise focus on an iris area$k$, which means the AquulaCam should keep the object $k$ in the best focus position $f^\text{best}_t$ during the movement of the object $k$ in a particular range at each time step $t$. The camera can quickly focus by adjusting the optical power of the liquid lens. Our algorithm needs to update the camera's focus position $f_t$ based on the previous image $I_{t-1}$ by

\begin{equation}
   f^\text{best}_t \leftarrow f_t = f_\text{t-1} + \bigtriangledown f_t
   \label{eq:update}
\end{equation}

In this section, we will discuss how to derive the focus-step $\bigtriangledown f_t$.

\subsection{Optical Lens System}

\begin{figure}[h]
   \begin{center}
      \includegraphics[width=0.8\linewidth]{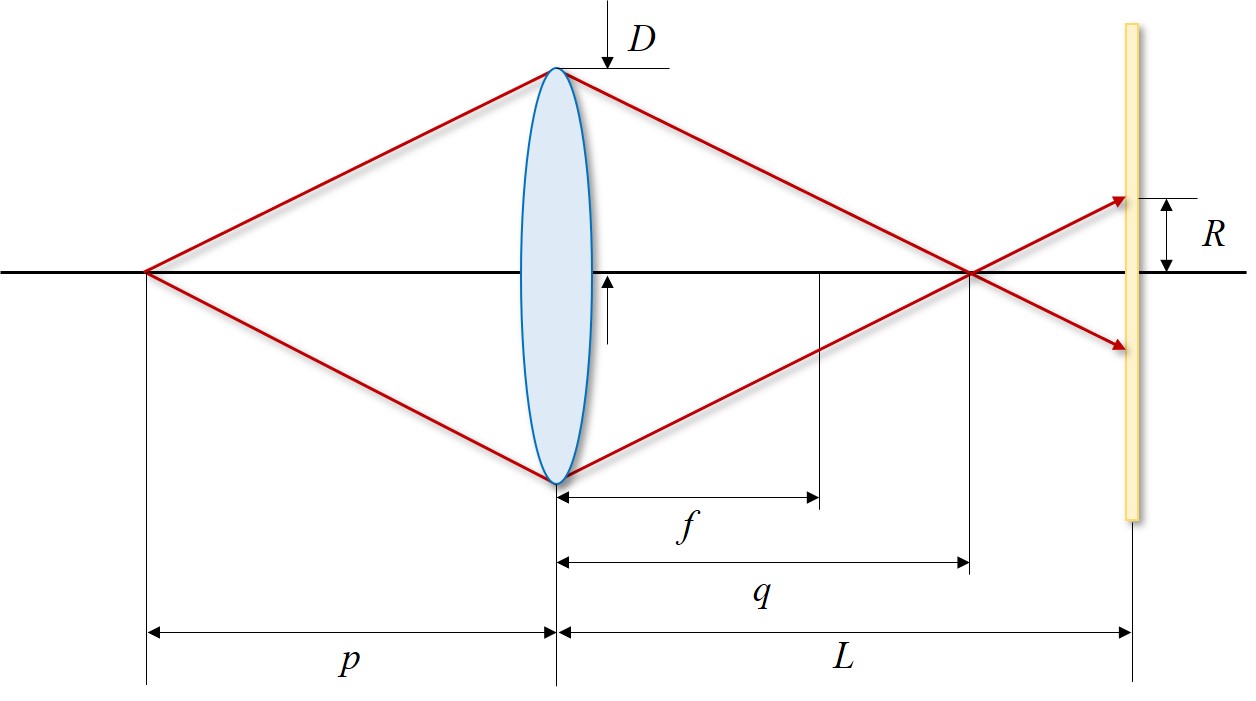}
   \end{center}
   \caption{Circle of confusion in a thin-lens optical system.}
   \label{fig:opt_sys}
\end{figure}

In the optical non-coherent imaging system, the imaging process follows the spatial translation invariance and satisfies linear systems' superposition principle. The imaging process of a complex object can be regarded as the convolution of a real object and point spread function (PSF). PSF is the response of the lens system for the point light source. The ideal PSF can be regarded as a circle if only considering defocus blur. However, the real lens system is not ideal, and practical PSF\cite{bracewell1995two} will be modified as

\begin{equation}
   h_{psf}(x,y) = \frac{1}{\sqrt{\pi R^2}}e^{-\frac{x^2+y^2}{R^2}}
   \label{eq:PSF}
\end{equation}

Where $R$ is the radius of the circle of confusion (CoC). CoC determines the depth of field (DOF), which is the distance between the nearest and the farthest objects in acceptably sharp focus in an image. Figure \ref{fig:opt_sys} demonstrates a simple thin lens system. According to geometric optics, the radius of CoC $R$ will be given by

\begin{equation}
   R = \frac{D}{q}|L-q|
   \label{eq:RDQ}
\end{equation}

Aperture radius, focal length, image plane location from the lens, and object distance image distance are represented as $D$, $f$, $L$, $p$, $q$. Considering that the $L$ of AquulaCam is fixed, $f$ is variable, and the thin lens equation

\begin{equation}
   \frac{1}{f} = \frac{1}{q}+\frac{1}{p}
   \label{eq:fpq}
\end{equation}

Using Eq.(\ref{eq:RDQ}) and Eq.(\ref{eq:fpq}), the relationship between $R$ and $f$ is

\begin{equation}
   R = \frac{D}{fp}+|Lp-f(L+p)|
   \label{eq:Rf}
\end{equation}

Based on above discussion, the relationship between PSF $h_{psf}(x,y)$ and focal length $f$ can be obtained from formulas Eq.(\ref{eq:PSF}) and Eq.(\ref{eq:Rf}). However, this is only a theoretical derivation, and the relationship between them in the real imaging system will be more complicated. But we can use deep neural networks to fit this function in AquulaCam.

In recent years, in the research of single image super-resolution\cite{guBlindSuperResolutionIterative2019} and deblurring\cite{xuMotionBlurKernel2018}, the use of deep learning to estimate PSF has achieved great success. Therefore, we believe that a deep neural network can be constructed to estimate the difference of focal length $\bigtriangledown f$ between the best focus position $f^\text{best}_t$ and the current focus position $f_\text{t-1}$ from the input image. We use MSE loss to regress $\bigtriangledown f_t$

\begin{equation}
   % L_\text{f} = \frac{1}{N} \text{L2Loss}(\bigtriangledown f, (f^\text{best}-f_\text{current}))
   L_\text{f} = \frac{1}{N} \sqrt{(\bigtriangledown f_t - (f^\text{best}_t-f_\text{t-1}))^2}
   \label{eq:L_f}
\end{equation}

\subsection{Object Detection in Blur Image}

In the previous section, we mentioned the superposition principle of imaging systems. If there are multiple objects in the scene that are not in the same focal plane, their PSFs are also different. Therefore, we need to know the position of the object we are tracking focus in the image. This is a typical object detection question\cite{liu2016ssd,girshick2015fast,lawCornerNetDetectingObjects2018}, but its difficulty lies in the fact that the iris area in the first frame of the image acquired by AquulaCam may be so blurry that humans cannot give an accurate bounding box\cite{girshick2015fast}. On the other hand, the bounding box cannot be used directly in the end-to-end network.

In this paper, we use semantic segmentation to localize the iris area. The value of the heatmap represents the probability that the pixel belongs to the iris area. $I \in R^{W \times H \times 3}$ is the input image of width $W$ and height $H$. We aim to produce a heatmap $Y \in [0,1]^{W \times H \times C}$, where $C$ is the number of categories. In training, we apply cross-entropy loss.

% For training, we apply the smooth L1 Loss at the heatmap $\hat{Y_{k}}$ similar to ground-truth Gaussian distribution $Y_{k}$

% \begin{equation}
%    L_\text{heatmap} = \frac{1}{N} \sum{z}
%    \label{eq:L_heatmap}
% \end{equation}
% where $z$ is given by

% \begin{equation}
%    z=
%    \begin{cases}
%       0.5(Y - \hat{Y})^2,  & if |Y - \hat{Y}|<1 \\
%       |Y - \hat{Y}| - 0.5, & otherwise
%    \end{cases}
%    \label{eq:zi}
% \end{equation}

\subsection{Network Architecture}

\begin{figure}[h]
   \begin{center}
      \includegraphics[width=0.95\linewidth]{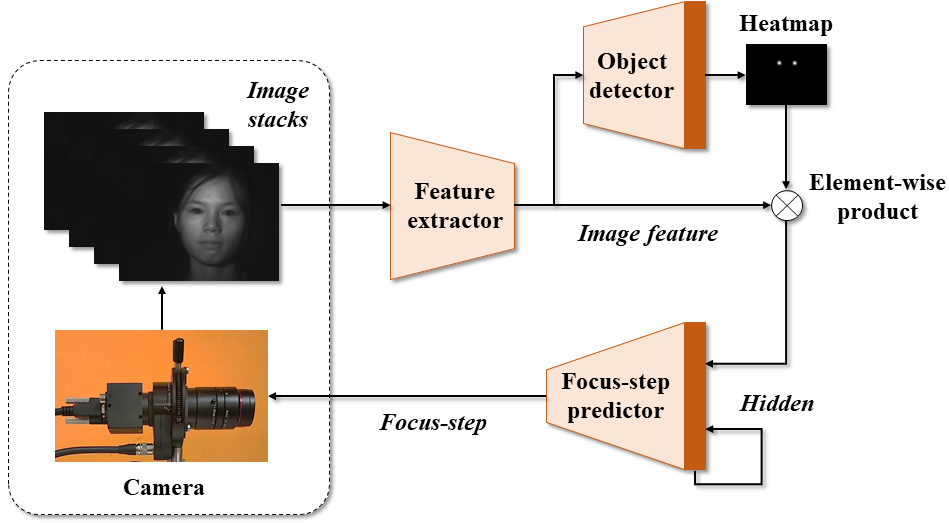}
   \end{center}
   \caption{An overview of the network architecture of AquulaCam.}
   \label{fig:network}
\end{figure}

The tracking focus network consists of three primary components: a feature extractor, an object detector, and a focus-step predictor shown as Figure \ref{fig:network}. The feature extractor aims to extract a feature map $\phi^t$ for the image captured at each time step $t$. The object detector produces the heatmap $\hat{Y_{k}^t}$. The focus-step predictor integrates these feature maps over time to derive a state representation $h^t$ at each time step $t$ and exploits the hidden state $h^t$ to obtain the focus-step $\bigtriangledown f$.

\noindent\textbf{Feature Extractor} The feature extractor transforms the image captured at each time step $t$ into a feature map $\phi^t$ as input to the object detector and focus-step predictor. In our network, we chose vanilla U-Net's encoder\cite{ronneberger2015u}, which includes three convolutional-pooling blocks, the number of channels is 16, 32, 64.

\noindent\textbf{Object Detector} As discussed in the previous section, the object detector produces the heatmap $\hat{Y_{k}^t}$, selects the channel corresponding to the category and resize it to the size of $\phi^t$. In this way, the attention map $\tau^t$ is obtained, which will be element-wise multiplied with the feature map $\phi^t$. We chose vanilla U-Net's decoder as an object detector.

\noindent\textbf{Focus-step Predictor} Focus-step Predictor includes a recurrent network and several fully connected layers. The element-wise product of feature map $\phi^t$ and attention map $\tau^t$ has reduced the dimensionality by global average pooling and produce the feature vector $\psi^t$. It only provides the information about \textit{what} and \textit{where} the object is, but not \textit{how} the object moves. Therefore, we used an LSTM to model the motion of the object. At one single time step $t$, $\psi^t$ and the hidden state at previous time step $h^{t-1}$, which implicitly memorizes the history capturing "how" the object moves, are fed in, and the renewed hidden state $h^t$ are returned. Finally, the fully connected layer uses $h^{t}$ to predict focus-step $\bigtriangledown f$. An LSTM and a single-layer, fully connected layer are used in our network.

\noindent\textbf{Loss Function} The overall training objective is
\begin{equation}
   L = L_{f} + \lambda L_{heatmap}
\end{equation}
We set $\lambda = 1$ in all our experiments unless specified otherwise.

%-------------------------------------------------------------------------
\section{Experiments}

\begin{figure*}[h]
   \begin{center}
      \includegraphics[width=0.95\linewidth]{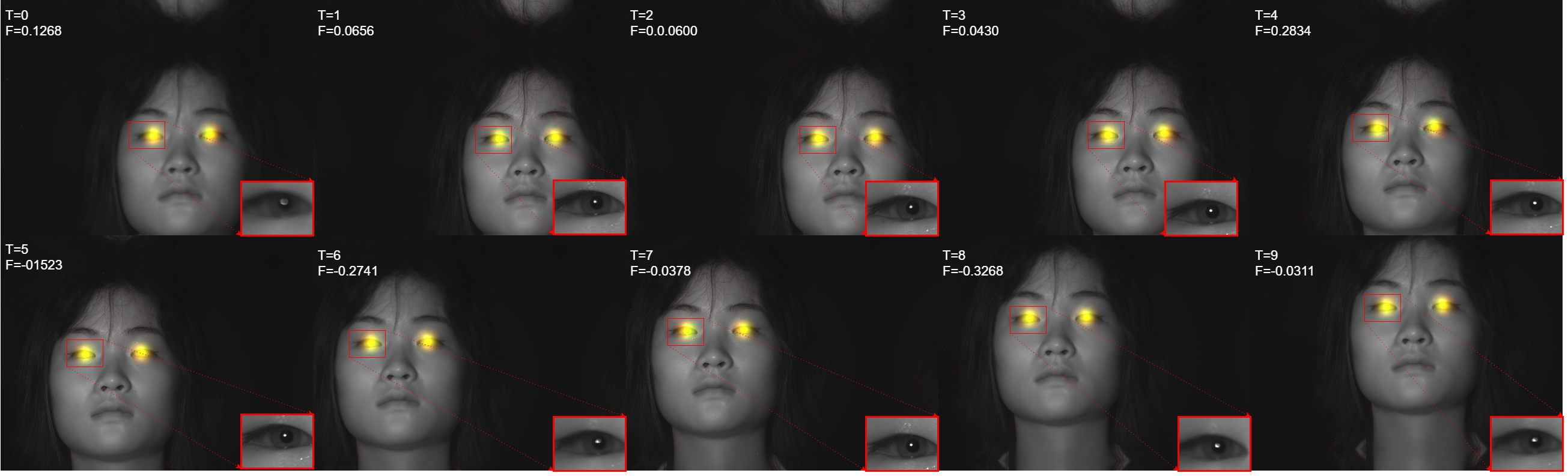}
   \end{center}
   \caption{\textbf{The result of focus tracking for a moving object.} AquulaCam is rapidly adjusted to the proper focus position starting from the second frame and manipulates all frames in focus afterward. The red layer is the predicted human eye area, the green layer is the ground-truth of the human eye area, and the yellow area is the overlap}
   \label{fig:track_img}
\end{figure*}

\subsection{Focus-Tunable Lens}
\label{sec:len}

% \begin{figure}[h]
%    \begin{center}
%       \includegraphics[width=0.8\linewidth]{img/opto_elec.png}
%    \end{center}
%    \caption{Typical data showing the relation between optical power (in diopters) and the control current. This picture is from the user manual of EL-16-40-TC\cite{Optotunelens}.}
%    \label{fig:opto_elec}
% \end{figure}

AquulaCam uses Optotune’s EL-16-40-TC (5D) electrically tunable lens\cite{Optotunelens}. Both the optical fluid and the membrane material are highly transparent at 400 to 2500 nm. By applying an electric current to this shape-changing polymer lens, its optical power is controlled within about 25 ms over a diopter range of -2 dpt to +3 dpt. An electromagnetic actuator is used to exert pressure on the container and therefore changes the curvature of the lens. By changing the electrical current flowing through the coil of the actuator, the optical power of the lens increases with a positive current and decreases with a negative current.

\subsection{Dataset and Virtual Camera}
\label{sec:dataset}

% \begin{figure*}[h]
%    \begin{center}
%       \includegraphics[width=0.95\linewidth]{img/move_demo.png}
%       \caption{Demonstration of a moving object simulation image sequence. The green area in the figure is the ground truth of the iris area.}
%    \end{center}
%    \label{fig:move_demo}
% \end{figure*}

% \begin{figure*}[h]
%    \begin{center}
%       \includegraphics[width=0.95\linewidth]{img/auto.png}
%    \end{center}
%    \caption{Performance of hill-climbing algorithm in stationary object autofocus. Although it might reach a better focus position within 4 iterations, most of the tests take more than 10 iterations.}
%    \label{fig:autofocus}
% \end{figure*}

We collected a dataset using the AquulaCam without autofocus, which was designed for long-distance uncontrolled iris recognition, so we collected images of human faces under near-infrared light. Volunteers were asked to stand and stay still, looking directly at the camera. AquulaCam traverses the current range permitted by the liquid lens and captures each current position's image to form a focus stack. There are 76 focus stacks in this dataset, and each focus stack contains about 80 images. Figure \ref{fig:focal_stack} is a part of the image in a focus stack.

In order to train the proposed network offline, we designed a virtual camera. Each virtual camera includes a focus stack, and each image in the focus stack has the current captured by AquulaCam. When the virtual camera is initialized, the first frame and the movement of the object perpendicular to the main optical axis (x-, y-axis) and along the main optical axis (z-axis) will be set. We simulate the movement of the object on the xy-axis by translating the image, and simulate the movement on the z-axis by adding an offset $f_\text{off}$ to the focus position. Specifically,

\begin{equation}
   I_\text{t} = \text{AquulaCam}(f^\text{t-1}+\bigtriangledown f_t+f^\text{off}_t)
\end{equation}

The virtual camera can simulate linear motion, swing, and random motion in three directions.

% We have realized the simulation of uniform linear motion, simple harmonic motion, and random motion in these three directions. The trajectory is as shown Figure \ref{fig:move}

% \begin{figure*}[h]
%   \begin{center}
%       \includegraphics[width=0.8\linewidth]{img/move_demo.png}
%   \end{center}
%   \caption{Simulated trajectory of a moving object.}
%   \label{fig:move}
% \end{figure*}

\subsection{Experimental Results}

We conducted a comparison experiment with the autofocus method based on the Fibonacci search\cite{gendlin1982focusing}. Specifically, there are static scenes and dynamic scenes, i.e., focusing on iris areas of fixed objects and moving objects. We assume the existence of an algorithm capable of localizing the iris area in blurred images and providing the ground truth as ROI to the Fibonacci search algorithm.

We mainly compare two metrics: the mean and standard deviation of the sharpness of the iris area $Y$ at each time step. The sharpness uses the Tenengrad criterion

\begin{equation}
   \text{T} = \frac{1}{\sum{Y}}\sum{Y\sqrt{(G_x*I)^2+(G_y*I)^2}}
\end{equation}

\noindent where

\begin{equation}
   G_x=
   \left[ \begin{matrix}
         +2 & 0 & -2 \\
         +4 & 0 & -4 \\
         +2 & 0 & -2
      \end{matrix}\right ],
   G_y=
   \left[ \begin{matrix}
         +2 & +4 & +2 \\
         0  & 0  & 0  \\
         -2 & +4 & -2
      \end{matrix}\right ]
\end{equation}

Considering the significant variation in Tenengrad values for different objects, we also normalize them between $[0,1]$.

The Fibonacci search algorithm is capable of achieving image processing speeds over 100fps, while our approach can also reach speeds of about 50fps. However, since the resolution of the original face image is 4080*3072, the speed of image acquisition is less than 25fps for both the real camera and the virtual camera. Therefore, we use the number of acquired images as the time metric in the experiment. The autofocus algorithm will reach a steady state after a certain number of iterations, so we only take the first 30 time step for comparison.

% \begin{figure}[h]
%   \centering
%   % \subfigure[Sharpness]{
%   %    \includegraphics[width=0.9\linewidth]{img/stay_fm.png}
%   %    \label{fig:stay_fm}
%   % }
%   % \subfigure[Focus position]{
%   %    \includegraphics[width=0.9\linewidth]{img/stay_ele.png}
%   %    \label{fig:stay_ele}
%   % }
%   \includegraphics[width=0.9\linewidth]{img/stay_fm.png}
%   \caption{Comparison of AF results in static scenes.}
%   \label{fig:stay_fm}
% \end{figure}

\begin{figure}[h]
   \begin{center}
      \subfigure[Static scenes]{
         \includegraphics[width=0.9\linewidth]{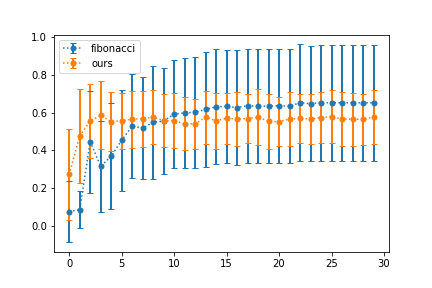}
         \label{fig:stay_fm}
      }
      \subfigure[Dynamic scenes]{
         \includegraphics[width=0.9\linewidth]{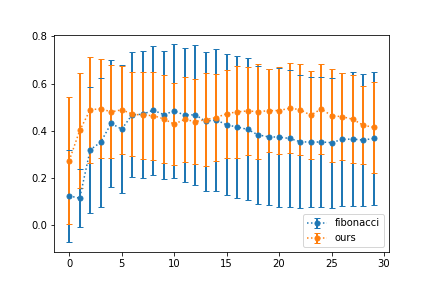}
         \label{fig:move_fm}
      }
   \end{center}
   \caption{Comparison of autofocus results, where the horizontal axis is the time step and the vertical axis is the normalized sharpness.}
   %   \label{fig:move_fm}
\end{figure}

\noindent\textbf{Static Scenes.} The first image acquired by our method is a randomly selected focus position, and it can be seen in figure \ref{fig:stay_fm} that our method achieves high sharpness on the second image and is able to maintain a relatively high sharpness after that. The Fibonacci search algorithm selects the images at either end of the focus position and gradually searches toward the middle, arriving near the best focus position at about 15 frames. The Fibonacci search algorithm has an average- and worst-case complexity of $O(log_2{n})$, consistent with the experimental results.
% The trend of the focus position in the figure \ref{fig:stay_ele} is the same as the figure \ref{fig:stay_fm}. Noted that our method has a larger range of focus position variation increased with iterations. This is due to the fact that our model is trained in a scene where the object is in motion and the model tries to compensate for the random motion of the object using some random vibrations, and this error is accumulated in the recurrent neural network. 
Comparing the two methods, our method has a faster focus speed but does not achieve the global best focus position. The Fibonacci search method, on the other hand, is bound to achieve the best focus position after multiple iterations, but the focus time may even exceed 600 ms, which much higher than the time required by other algorithms in iris recognition systems.

% \begin{figure}[h]
%   \begin{center}
%       % \subfigure[Sharpness]{
%       %    \includegraphics[width=0.9\linewidth]{img/move_fm.png}
%       %    \label{fig:move_fm}
%       % }
%       % \subfigure[Focus position]{
%       %    \includegraphics[width=0.9\linewidth]{img/move_ele.png}
%       %    \label{fig:move_ele}
%       % }
%       \includegraphics[width=0.9\linewidth]{img/move_fm.png}
%   \end{center}
%   \caption{Comparison of AF results in dynamic scenes.}
%   \label{fig:move_fm}
% \end{figure}

\noindent\textbf{Dynamic Scenes.} As we discussed before, all image-based autofocus algorithms search for the maximum point of the sharpness curve within a given interval. And when the object moves, its sharpness curve also moves. Therefore this type of algorithm cannot focus on a moving object. This can be seen in Figure \ref{fig:move_fm}, where the sharpness of the iris area fluctuates considerably with the movement of the object. In contrast, our method is able to model the motion of the target within two time steps and maintain a relatively ideal focus. Figure \ref{fig:track_img} shows the images acquired under the control of our method, which allows AquulaCam to acquire clear enough iris images for recognition consistently.

\subsection{Demonstration in Real-world Scenarios}

The test in the real-world scenario is similar to the results shown in the previous section. Since it is impossible to label the moving objects in real-time manually, we did not perform quantitative tests. A demonstration in a real-world scenario can be found in the supplemental material(https://disk.yandex.com/d/MPSdU4dIkRmrRA). AquulaCam and the near-infrared light source were placed on the table, and the volunteers were sitting in front of a table, moving randomly. Our model has achieved a speed of over 50 fps on the Nvidia Quadro K620, which is higher than the zoom speed of the lens 40 fps and the capture speed of the image sensor 15 fps. It is demonstrated this AF camera is able to track the ideal focus position in real-time and can maintain the ideal focus for more than 200 frames.

\begin{figure}[h]
   \begin{center}
      \includegraphics[width=0.85\linewidth]{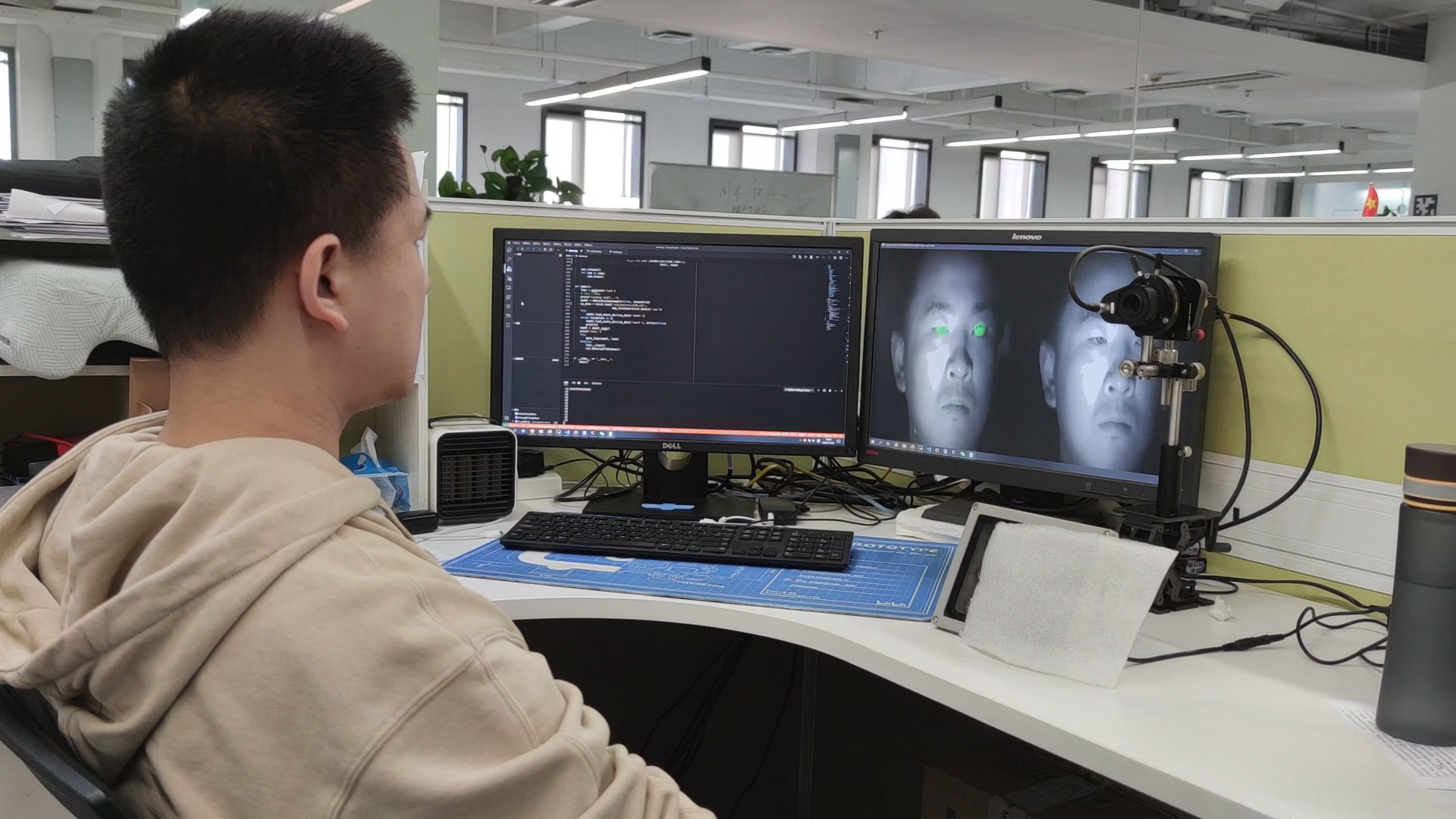}
   \end{center}
   \caption{Demonstration in real-world.}
   \label{fig:demo}
\end{figure}
% At present, we have achieved eye tracking focus under near-infrared light. The more complex background and different types of objects in the visible light scene will bring greater challenges, which is the direction of our future work. On the other hand, the bottleneck of focusing speed is our image sensor. If we upgrade to a higher capture speed image sensor, the speed of the system should be greatly increased. In addition, we have considered use the piezo element control tunable-lens to replace this Optotune VCM control lens for even faster response time and less power consumption. So it may be possible to use time division multiplexing to maintain well focus positions of multiple moving objects at the same time.

%------------------------------------------------------------------------
\section{Conclusion}

In the past decade,  iris recognition gradually develops from close controlled scenes to uncontrolled long-distance scenes. Imaging systems need to quickly focus on objects in different positions and maintain focus on a moving user's eyes. In this paper, we introduced a novel rapid autofocus camera, AquulaCam, to actively visualize the iris area of moving objects using the focus-tunable lens. We demonstrate that the end-to-end computational algorithm can actively perceive the iris area and generate a lens diopter control signal from one blurred image. Our algorithm's speed and the ability to model motion enable real-time focus tracking of a moving object. We built a testing bench to collect real-world focal stacks for evaluation of the autofocus methods. The results demonstrate the advantages of AquulaCam for biometric perception in static and dynamic scenes. It is expected to extend this autofocus camera to the areas of active perception, continuous tracking, and precise focus control.

%------------------------------------------------------------------------
~\\
\noindent \textbf{Acknowledgments} The authors would like to thank the anonymous reviewers for their valuable comments and advices. This work is supported by National Natural Science Foundation of China (Grant No. 62071468, 62006225, 61806197, 6207146) and Strategic Priority Research Program of Chinese Academy of Sciences (Grant No.XDA27040700).

   %------------------------------------------------------------------------
   {\small
      \bibliographystyle{ieee}
      \bibliography{mybib}
   }

\end{document}